\documentclass[letterpaper, 10 pt, conference]{ieeeconf}  

\IEEEoverridecommandlockouts                              

\usepackage{mathptmx} 
\usepackage{amsmath} 
\usepackage{dirtytalk}
\usepackage{graphicx}
\usepackage{array}
\usepackage{booktabs}
\usepackage{color}

\graphicspath{ {images/} }

\title{\LARGE \bf
Toward Autonomous Robotic Micro-Suturing using Optical Coherence Tomography Calibration and Path Planning
}

\author{Yuan Tian$^1$, Mark Draelos, PhD$^2$, Gao Tang$^3$, Ruobing Qian$^2$,\\ Anthony Kuo, MD$^4$, Joseph Izatt, PhD$^2$, Kris Hauser, PhD$^3$
\thanks{This work is partially supported by NIH Grant R21-EY029877.}
\thanks{$^1$ Y. Tian is with the Department of Electrical and Computer Engineering, Duke University, Durham, NC, USA. {\tt\small yuan.tian277@duke.edu}}%
\thanks{$^2$ M. Draelos, R. Qian and J. Izatt are with the Department of Biomedical Engineering, Duke University, Durham, NC, USA.}%
\thanks{$^3$ G. Tang and K. Hauser are with the Department of Computer Science, University of Illinois Urbana-Champaign, Urbana, IL, USA.}%
\thanks{$^4$ A. Kuo is with the Department of Ophthalmology, Duke University Medical Center, Durham, NC, USA.}%
}

\begin{document}

\maketitle
\thispagestyle{empty}
\pagestyle{empty}

\begin{abstract}

Robotic automation has the potential to assist human surgeons in performing suturing tasks in microsurgery, and in order to do so a robot must be able to guide a needle with sub-millimeter precision through soft tissue. This paper presents a robotic suturing system that uses 3D optical coherence tomography (OCT) system for imaging feedback.  Calibration of the robot-OCT and robot-needle transforms, wound detection, keypoint identification, and path planning are all performed automatically. The calibration method handles pose uncertainty when the needle is grasped using a variant of iterative closest points. The path planner uses the identified wound shape to calculate needle entry and exit points to yield an evenly-matched wound shape after closure. Experiments on tissue phantoms and animal tissue demonstrate that the system can pass a suture needle through wounds with 0.200\,mm overall accuracy in achieving the planned entry and exit points, and over 20$\times$ more precise than prior autonomous suturing robots. 

\end{abstract}

\section{INTRODUCTION}

Suturing is a basic surgical skill used in microsurgery to repair wounds and severed blood vessels and nerves, but it is tedious, time-consuming, and requires substantial training~\cite{alrasheed2014robotic}.  Robotic assistance has been proposed for suturing in open surgery, laparoscopic surgery, and microsurgery, with teleoperated surgical systems increasing the surgeon's dexterity, as well as autonomous techniques that can alleviate burden on the surgeon.  Automation of suturing is still a challenging task due to deformation of the tissue as the needle passes through it, the need for regrasping to complete multiple throws, and incorporation of imaging feedback to estimate wound shape and guide the needle successfully along planned paths.  Tracking of needles and 3D tissue is challenging due to translucent tissue, deformation, and occlusion of the needle as it passes through the tissue.  The challenge is particularly acute in the microsurgery setting, since existing sensing and actuation technologies do not perform at the required level of precision needed to complete reliable suturing.

\begin{figure}[tb]
    \centering
    \includegraphics[scale=0.45]{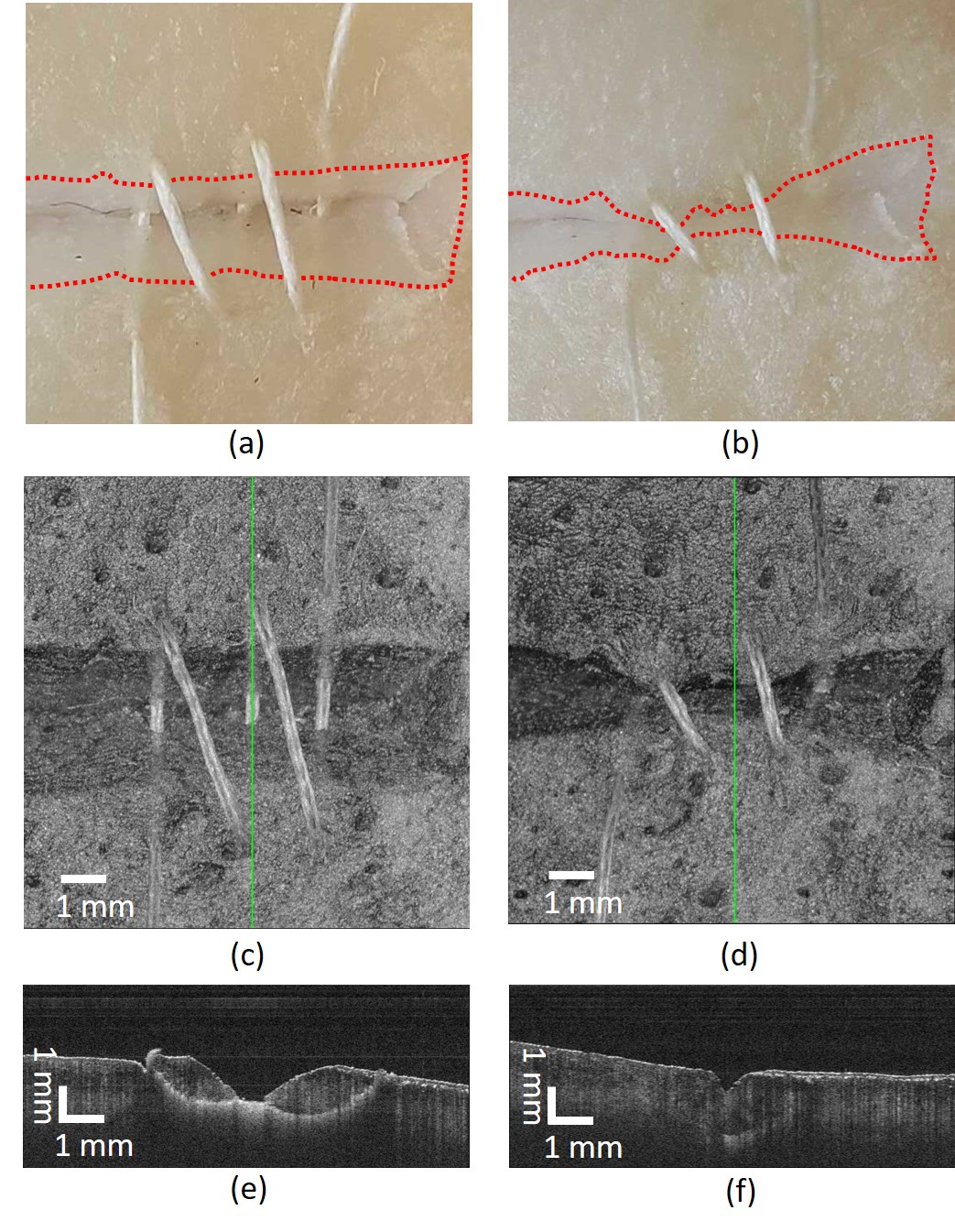}
    \caption{Our system performs a sub-centimeter multi-throw suture in porcine tissue with a notch-shaped wound. Dotted lines delineate wound edges before (a) and after (b) closure, showing good approximation between the second throw's right bite point and left exit point. The OCT volume projection is shown before (c) and after (d) closure. (e) The highlighted B-scan (green line in (c)) in the OCT volume before closure. (f) The B-scan after closure, showing no dead space \cite{Ethicon2007}.}
    \label{fig:final_suture}
\end{figure}

\begin{figure}[t]
    \centering
    \includegraphics[scale=0.45]{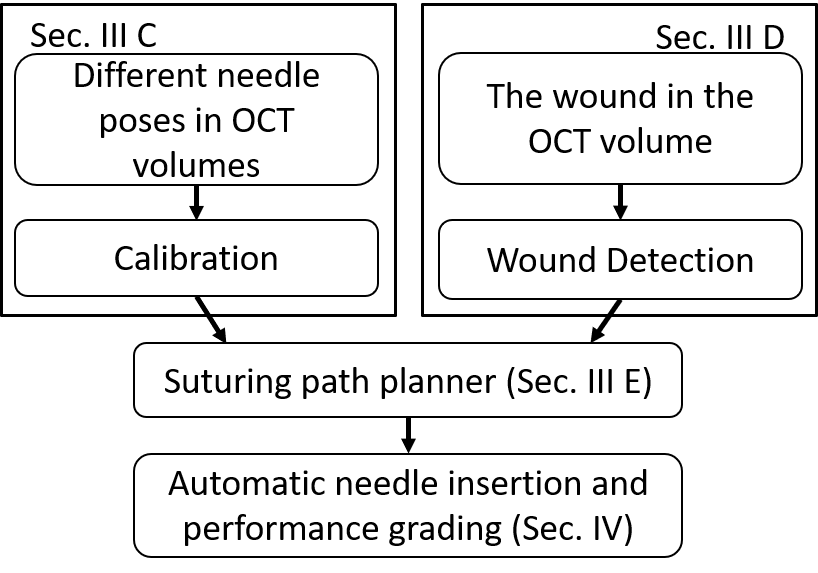}
    \caption{The flow chart of the overall robotic suturing procedure, which can be divided into coordinates calibration, wound detection, suturing path planning,  automatic needle insertion, and performance grading.} 
    \label{fig:flowchart}
\end{figure}

This paper presents a robotic system that achieves sub-millimeter suturing precision, consisting needle calibration, planning, and insertion under optical coherence tomography (OCT) imaging guidance (Fig.~\ref{fig:final_suture}). OCT is a laser-based imaging technique that exploits interference to resolve the distance of reflectors along the beam's path to within several micrometers \cite{OCT}. By scanning the beam, a cross-sectional view (or B-scan) of the tissue is produced, and further sweeping the beam in a raster pattern yields a 3D volume (a $\sim$1\,cm$^3$ cube in our system).  Although OCT has deepest penetration in transparent tissues like those of the eye, it can also be used to image the surface and first millimeters in depth of scattering tissues like skin.  OCT has also been used to track tissues and surgical tools in eye surgery~\cite{Keller2018}. To guide the needle precisely, we perform calibration, perception, and planning in the OCT imaging frame. To calibrate the needle tip and robot with respect to the OCT frame, we use the observed needle surface voxels to simultaneously optimize the needle, robot, and OCT frames with a modified iterative closest points (ICP)~\cite{ICP} algorithm that emphasizes an accurate fit at the needle tip. A suturing path planner chooses a suturing path such that the two sides of the wound will match after closure, without dead space~\cite{Ethicon2007}.  Fig.~\ref{fig:flowchart} shows the overall procedure for our robotic suturing process.

Closed-loop imaging guidance is challenging due to OCT artifacts including refraction and shadowing by the needle driver.  Instead, our experiments observe that open-loop execution of a circular suture path leads to high accuracy between the planned and actual needle entry and exit points, even under significant tissue deformation. A unique aspect of our experiments is that we use 3D OCT to image the resulting suture path after the needle is pulled through and the tissue returns to a rest state. Experiments in a tissue phantom and porcine skin show the targeted entry and exit points are reached with root mean square error (RMSE) of 0.200\,mm, which is less than half the width of the needle and over 20$\times$ more precise than prior autonomous suturing robots.

\section{RELATED WORK}

Several researchers have studied the use of robots and imaging guidance for suturing and micromanipulation assistance.  Using direct teleoperation, human surgeons are able to complete suturing for anastomoses of 3mm artificial blood vessels using the Intuitive Surgical DaVinci robot~\cite{alrasheed2014robotic}, but to our knowledge surgical robots have not yet been used for microsurgical suturing in vivo.

Autonomous suturing has been studied for at least two decades.  Early work by Kang and Wen proposes a laparoscopic robot that can perform manually controlled or semi-autonomous motions for several tasks, including suturing and automatic knot tying~\cite{Kang2000}. Each motion is initiated by a surgeon.  Staub et al. present an image-based guidance system for improved accuracy in guiding a suturing needle to pierce the tissue at a surgeon-indicated spot, as directed by a laser pointer~\cite{Staub2010}. 3--10\,mm errors were observed from their system.
More recent work has addressed multi-throw suturing using a dual armed laparascopic robot, using a special needle gripper that reduces grasp positioning errors and feedback from a stereo vision system~\cite{Sen2016}.  Their calibration system obtains needle predictions with translational error of approximately 2.9mm, and completes 86\% of throws. 
Our path planner is highly related to the method of Jackson and Cavusoglu for path planning of needles entering triangular shaped wounds~\cite{Jackson2013}. The geometry of a circular needle entering a wound was analyzed, and we adopt many of the same conventions here. Their paths were executed on a DaVinci robot with a 12.7\,mm radius needle.  A similar approach was adopted by Pedram et al. who studied the problem of needle selection for varying wound geometries~\cite{Pedram2017}. The errors of their entry and exit points were greater than 4.5\,mm with a 15\,mm radius needle.  Overall, accuracy must be improved significantly for autonomous suturing in microsurgery.  Our system integrates the planner with imaging feedback, calibration, and analysis of errors in the suture path for a needle with 4.14\,mm radius, and we obtain over an order of magnitude improvement in accuracy.

In the microsurgical domain, OCT-guided robots have been studied as a method for assisting surgeons to complete telemanipulated tasks, primarily in the eye. Yu et al. present a B-mode OCT-integrated forceps tool for haptic-controlled microsurgery to assist in  retinal membrane peeling~\cite{Yu2015}. Nasseri et al. present an OCT and robot guidance system to assist in precise injections for macular degeneration~\cite{Nasseri2017}. Draelos et al. present a hand-guided robot that provides stabilization and OCT guidance in the cornea \cite{Draelos2018}, which they further extend with autonomous needle insertion capabilities~\cite{Draelos2019}.

Our calibration process is highly related to the work of Zhou et al. that presents an OCT-based needle tip tracking and calibration scheme~\cite{Zhou2018}. This work identified pixels likely to belong to the needle from the OCT B-scans and segmented the needle from the background using a voting scheme.  Calibrating the robot-OCT frame and tracking the needle led to an impressive $\approx$10\,$\mu$m error as the needle moved along an XYZ translation stage. They also extended their method to 6 degree of freedom (DOF) tracking using ICP with similar levels of accuracy~\cite{Zhou2019}. Our work differs in that we consider a curved needle grasped by a needle driver, which makes needle identification more difficult due to artifacts caused by shadowing, saturation artifacts, and mirror images from the complex conjugate of the needle driver. Furthermore, our calibration process accounts for regrasping errors rather than using a fixed needle mount.


\begin{figure}[b]
    \centering
    \includegraphics[scale=0.32]{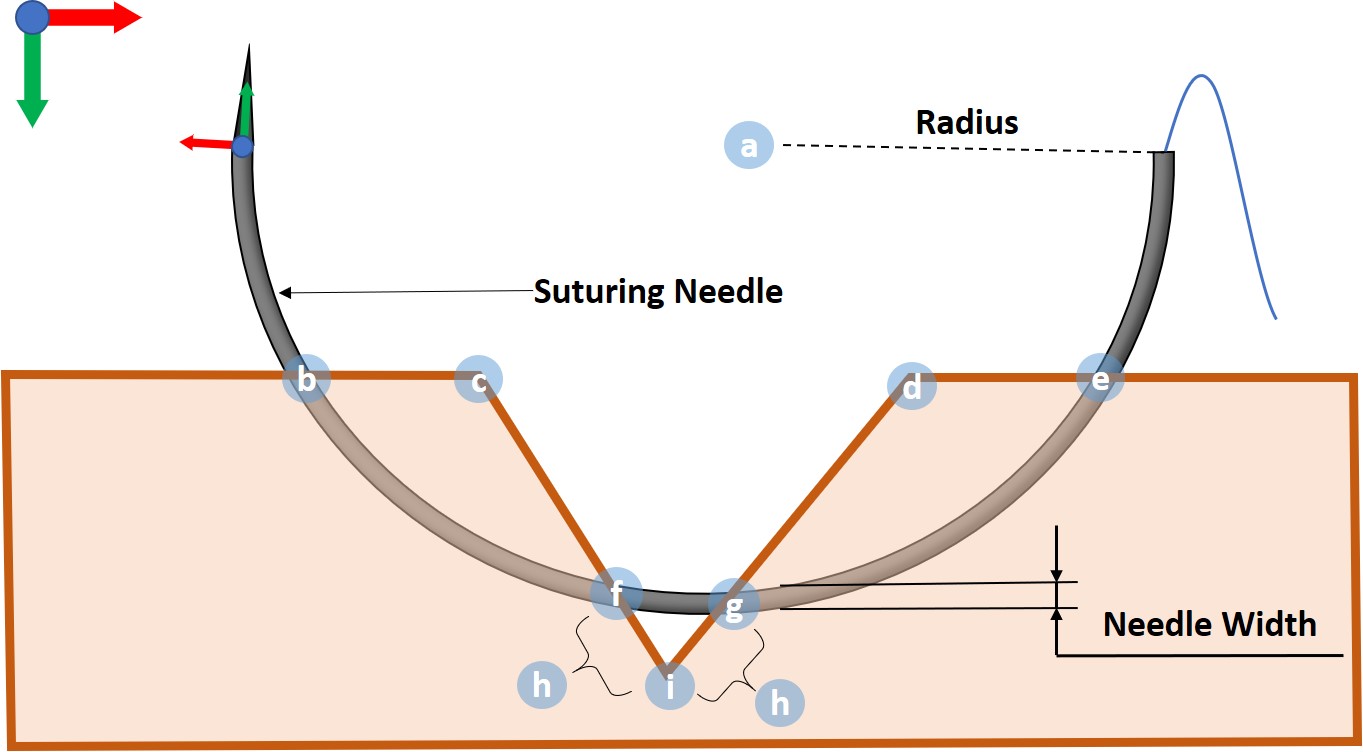}
    \caption{Cross-sectional geometry of the incised wound and suturing path, showing a) suturing center, b) left exit point, c) wound start point, d) wound end point, e) right bite point, f) left bite point, g) right exit point, h) distance from the deepest point to the left bite point (set to be equal to the distance to the right exit point), i) wound deepest point. } 
    \label{fig:woundshape}
\end{figure}

\begin{figure}[b]
    \centering
    \includegraphics[scale=0.31]{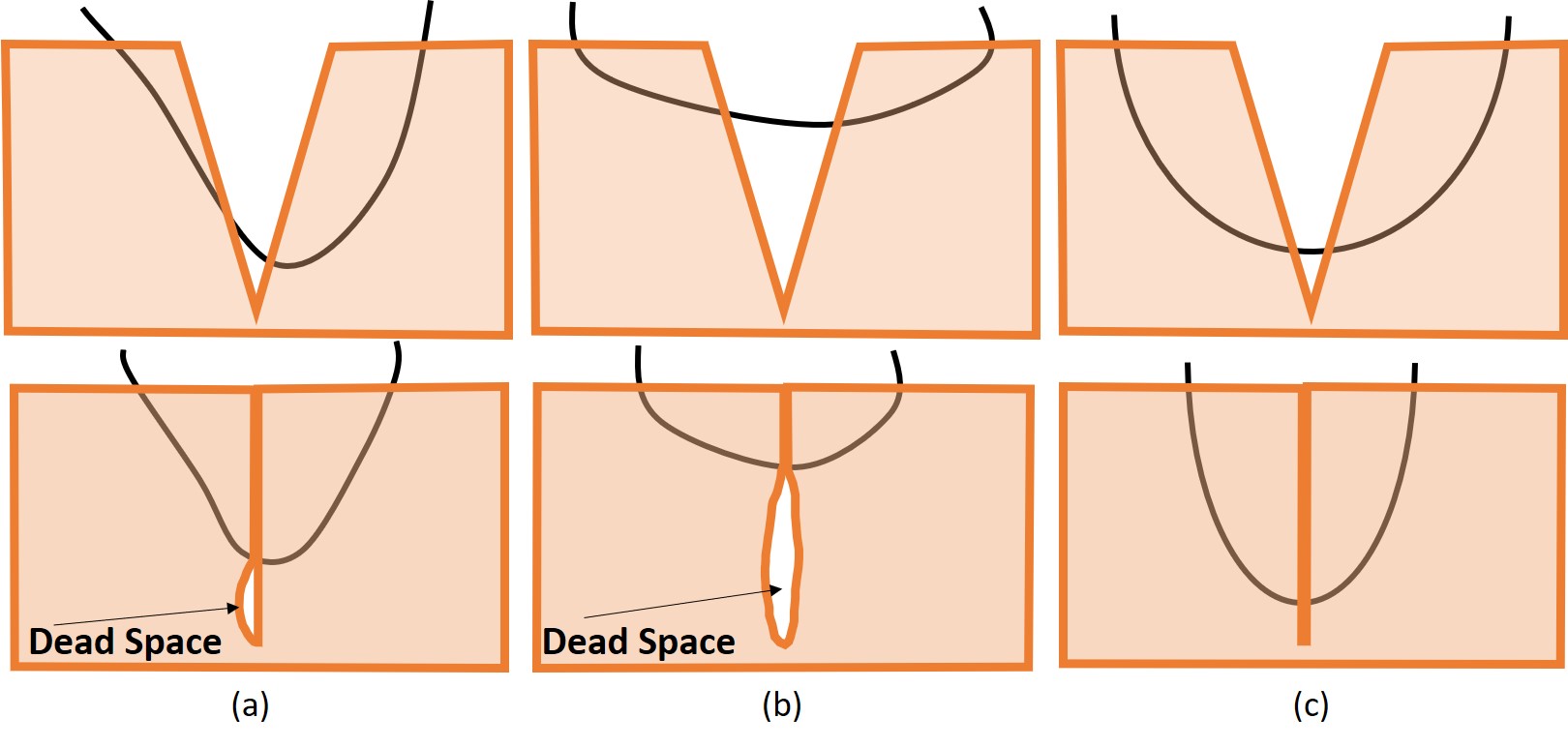}
    \caption{(a) The left bite point is higher than the  right exit point, so a dead space remains after closure. (b) The suturing depth is not sufficiently deep, which leaves a dead space after the wound is closed. (c) When the suturing path equalizes the wound sides and is sufficiently deep, then no dead space remains post-closure. }
    \label{fig:woundclose}
\end{figure}

\section{METHODS}

Our system is designed to calibrate a robot-mounted circular needle and automatically insert it through a wound, all using OCT imaging. To focus on the calibration and path planning problems, we show only the single-throw problem and leave multi-throw suturing for future work.  Our method assumes that 1) the needle body is a circular arc; 2) the needle diameter is larger than the wound width, so that no significant travel is needed to bring the wound sides together; 3) the entire wound is within the robot and OCT workspaces; and 4) the patient anatomy permits needle insertion by rotation about an axis parallel to the wound.

Consider the plane of the needle and a cross sectional view of a wound, where the needle is passing from right to left. To close the wound, the needle passes through the right bite point, right exit point, left bite point, and the left exit point in sequence (Fig. \ref{fig:woundshape}).  If the right exit point and left bite point are not well-aligned with the wound's deepest point, the resulting dead space after wound closure (Fig. \ref{fig:woundclose}) promotes infection~\cite{Ethicon2007}.  Our planner generates a suture path that avoids such dead space.

\subsection{Error Analysis}

The overall needle guidance error is influenced by many elements:
\begin{itemize}
    \item {\em Needle grasping uncertainty} is introduced when the needle driver grasps the needle.
    \item {\em Wound calculation errors} can be introduced during wound segmentation and keypoint identification.
    \item {\em Needle and OCT calibration error} remain after calibration, although needle grasping uncertainty is reduced. 
    \item {\em Robot repeatability and accuracy} limits the overall system performance. 
    \item {\em Tissue deformation} from friction as the needle passes through the tissue can be significant.
    \item {\em Conjugate image} overlap in OCT can produce artifacts that cause errors in needle identification. 
\end{itemize} 

We mitigate these errors through calibration to significantly reduce the needle grasping uncertainty and surgeon supervision of wound analysis. Robot error is minimized with a repeatability of 10\,$\mathrm{\mu m}$. The overall error is thus dominated by residual calibration error and tissue deformation. Closed-loop corrections under OCT imaging feedback would further reduce this error, but mid-insertion images suffer from shadowing, saturation artifacts, and refraction.

\begin{figure}[b]
    \centering
    \includegraphics[scale=0.30]{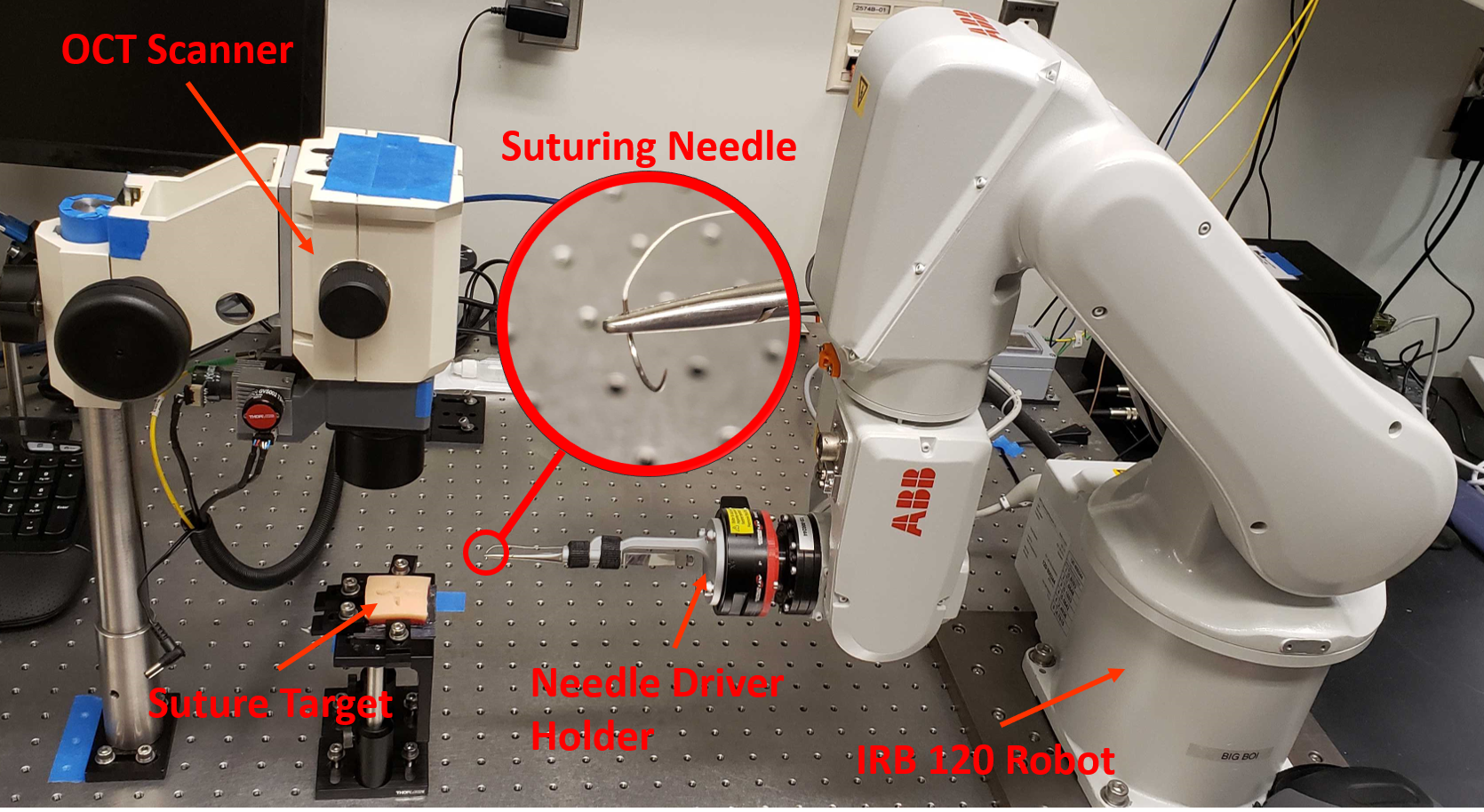}
    \caption{Photo of the experimental hardware.  On the left are the OCT scanner and suture target. On the right, the IRB 120 robot holds a needle driver holder, which holds a needle driver, which holds a suturing needle.}
    \label{fig:hardware}
\end{figure}

\subsection{Hardware}

We use a custom OCT engine with a 100 kHz swept-source laser centered at 1060\,nm (Axsun Technologies; Billerica, MA), an adjustable transmissive reference arm, and balanced detection. We configured this engine~\cite{Draelos2018} to capture OCT volumes with sampling density $725 \times 800 \times 1327$ pixels and field of view $10 \times 10 \times 7.15$ mm.
Our robot to perform the suturing needle insertion is a 6-joint IRB 120 Robot (ABB Robotics; Shanghai, China) with a specified repeatability of 10\,$\mu$m. 

A taper point, half-circle 4-0 suturing needle with 13\,mm arc length is used for tissue phantom and porcine skin insertions. The needle's width, as marked in Fig.~\ref{fig:woundshape} is 0.448\,mm. The needle is grasped with a locking Castroviejo needle driver (Ambler Surgical; Exton, PA) which is mounted to the robot's end-effector with a custom holder.
This holder keeps the grasped needle's rotation axis approximately coincident with the robot's last joint axis. This minimizes robot wrist and elbow movement to avoid collision with the surgical field and patient. The experimental hardware is shown in Fig.~\ref{fig:hardware}.

\subsection{Calibration}

Our calibration procedure calculates the transformations from the world frame to the OCT frame and from the robot end-effector frame to the needle frame.  
We define the needle origin at the interface between the needle tip and the needle body which facilitates path planning. The needle tip's conical nature causes the body and tip to follow separate paths when rotated~\cite{Jackson2013}. This misalignment may cause friction and deformation between the needle body and the suturing path.

\begin{figure}[b]
    \centering
    \includegraphics[scale=0.32]{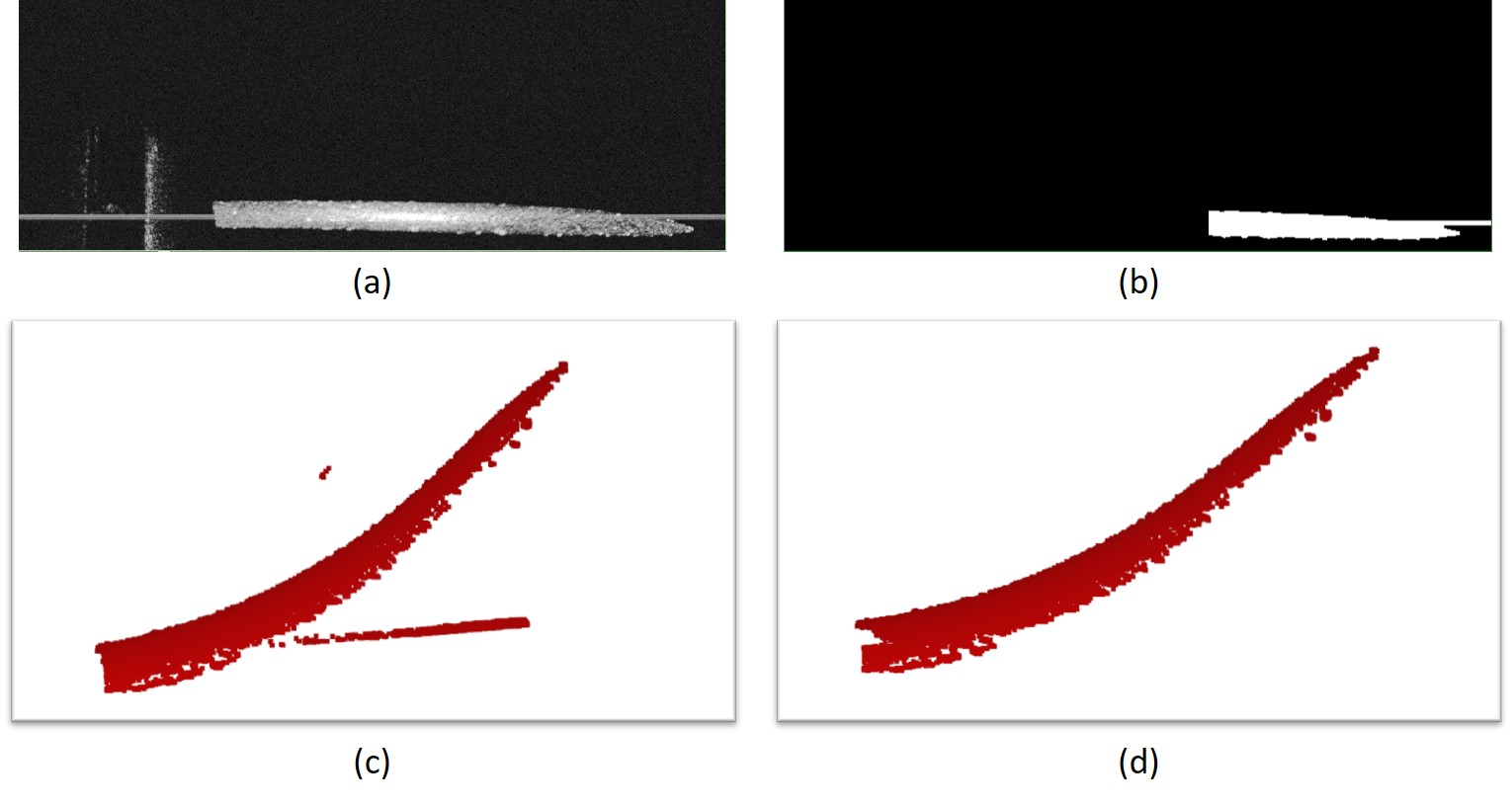}
    \caption{(a) The clipped projection of an OCT volume during calibration. (b) The largest connected component after clipping out the needle driver. (c) The point cloud including the part of needle, the saturation artifacts, and the noise. (d) The point cloud with saturation artifacts and noise removed.}
    \label{fig:needle_seg}
\end{figure}

\subsubsection{Needle Segmentation in OCT volumes}

Our needle segmentation method is similar to \cite{Keller2018}. We first generate the projection map of the OCT volume (Fig. \ref{fig:needle_seg}(a)), and apply a brightness threshold to obtain a binary image. As the needle driver's projection always appears at one side of the binary image, we clip the corresponding side of the image to remove the needle driver. We then apply the connected component algorithm \cite{BBDT} to find the largest connected component in the volume, which includes parts of the needle and the saturation artifacts (Fig. \ref{fig:needle_seg}(b)). 
The points in the OCT volume under the connected component's projection are the needle and the saturation artifacts' point clouds with noise (Fig. \ref{fig:needle_seg}(c)).
We then remove the saturation artifacts in the OCT volume, which appear as straight lines that are parallel to the OCT volume's B-scan direction and always touch or approach the edge in the OCT volume. Finally, we perform point cloud outlier removal \cite{Open3D} to filter out the noise below and above the needle's point cloud (Fig. \ref{fig:needle_seg}(d)).

\subsubsection{ICP Estimates of Needle Tip Transform}

\begin{figure}[t]
    \centering
    \includegraphics[scale=0.32]{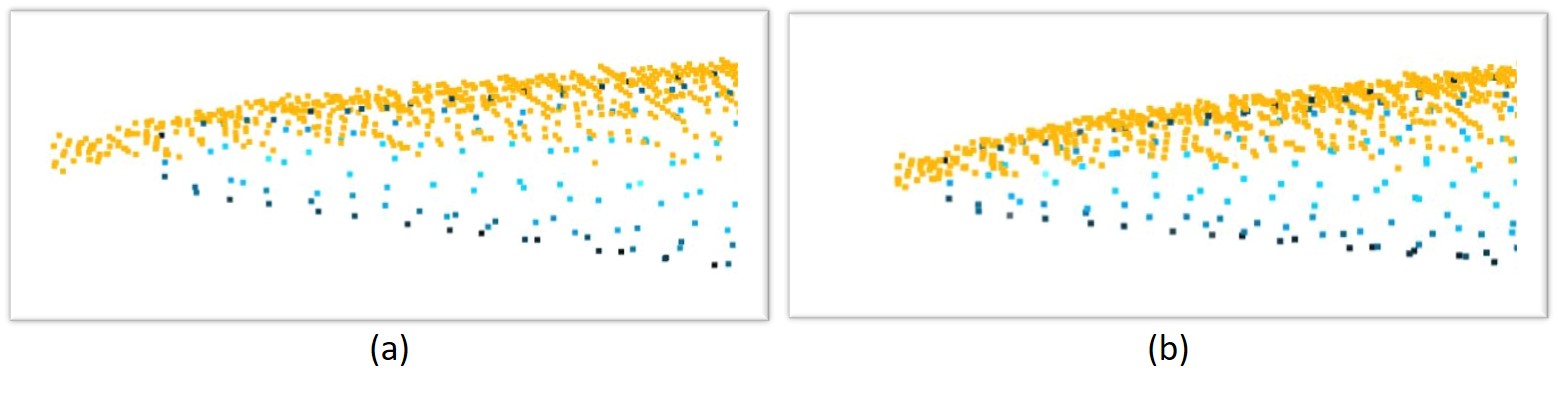}
    \caption{The yellow point clouds are the source point clouds captured in the OCT system, and the blue point clouds are target point clouds built in the CAD software. (a) Original ICP result. We transfer the source point cloud to the target point cloud frame. The needle tip of these 2 point clouds has mismatching. (b) The modified ICP result. The mismatching of the 2 point clouds has been alleviated. (Best viewed in color)}
    \label{fig:compared_icp}
\end{figure}

To estimate the needle transform in the OCT frame, we use Iterative Closest Points~\cite{ICP} which aligns points in the OCT field (source point cloud) and the needle CAD model (target point cloud). We identify the needle points as a subset of the bright pixels in the OCT maximum intensity projection (MIP). 
Standard ICP frequently mismatches the needle tip, however, because the point is dominated by body points (Fig.~\ref{fig:compared_icp}(a)). We overcome this problem by estimating tip points from an initial ICP fit and increasing their weight for a second ICP fit (Fig.~\ref{fig:compared_icp}(b)).
The second fit yields better alignment, especially at the needle tip.

\subsubsection{Calibration Algorithm}
We seek to solve
\begin{equation}
T_{ICP} = T_{OCT}^{-1}T_{EE}T_{N},
\label{eq:calibration}
\end{equation}
where $T_N$ is the needle frame in the end-effector frame, $T_{OCT}$ is the OCT frame in the world, $T_{EE}$ is the end-effector frame in the world from forward kinematics, and $T_{ICP}$ is the ICP-derived transformation from the needle frame to the OCT frame.
As described below, the robot will be guided to $m$ different poses, yielding paired observations $T_{ICP_i}$ and $T_{EE_i}$, for $i=1,\ldots,m$. 
We minimize the error in Eq.~\ref{eq:calibration} over $T_{N}$ and $T_{OCT}$ using a Levenberg-Marquardt method. 
%

\subsubsection{Calibration Set Sampling}

Our calibration set is an automatically defined set of $m=9$ robot configurations.  These should be chosen so that needle poses are sampled roughly uniformly across the OCT workspace and with a diversity of needle orientations. The needle tip's point cloud must also be visible and not shadowed by the body or needle driver. 

First, we manually jog the robot to move the needle tip into the center of the OCT volume. The robot then moves its end effector a small distance to produce two additional needle poses. Based on these 3 poses, we perform a rough calibration of $\tilde{T}_{OCT}$ and $\tilde{T}_{N}$. Because these poses are near each other and the OCT volume's center, the rough calibration is typically not sufficiently precise at the edge of the OCT frame. We thus define $m$ desired ICP poses $T_{ICPdes}$ in the OCT frame such that the needle's poses uniformly sample the field of view with different orientations. Using the below equation
\begin{equation}
T_{EE} = \tilde{T}_{OCT}T_{ICPdes}^{-1}\tilde{T}_{N}^{-1},
\label{eq:auto_sample}
\end{equation}
we compute $m$ end effector transforms, and $m$ robot configurations are automatically generated by inverse kinematics (IK) solver~\cite{Klampt}. Calibration is run a final time to obtain more precise estimates of $T_{OCT}$ and $T_{N}$.

\begin{figure}[t]
    \centering
    \includegraphics[scale=0.5]{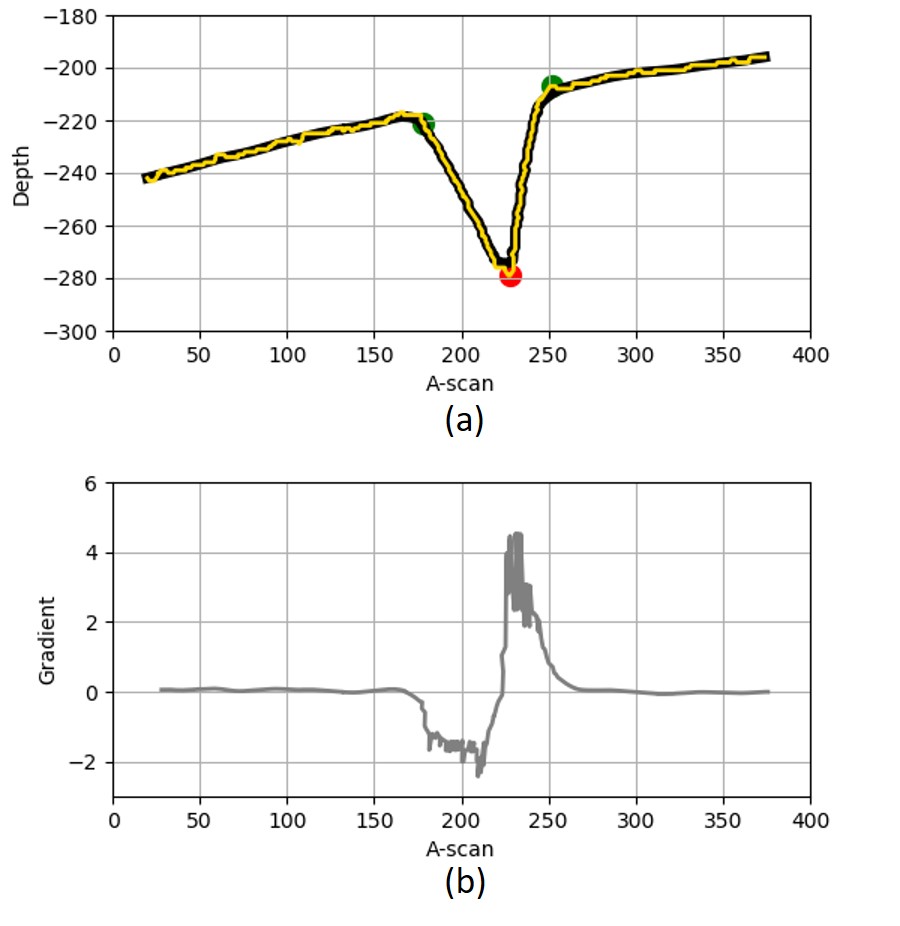}
    \caption{Wound top layer detection and geometric analysis. (a) The AMAL detection result is shown in yellow while the de-noised reult is shown in black. The estimated start and end of the wound are shown in green, while the deepest point is shown in red. (b) The gradient of the top layer of the wound after being de-noised.}
    \label{fig:wound_ana}
\end{figure}

\subsection{Wound Analysis}

Wound analysis processes an OCT B-scan image and produces the following shape information: the top layer, the start point, the end point and the deepest point.  We begin by using the adjusted mean arc length (AMAL) graph search method of Keller et al.~\cite{AMAL} to find the top layer of the wound (Fig.~\ref{fig:wound_ana}(a)). Next, we rotate the top layer such that the non-wound portion is horizontal. We use a Gaussian 1D filter to denoise the top layer, and then calculate the gradient of the smoothed top layer (Fig. \ref{fig:wound_ana}(b)).  We define the start point as the first point from left to right where the gradient falls below a threshold. Similarly, we define the end point as the first point from right to left where the gradient exceeds a threshold. The point that has smallest position is the deepest point of the wound.

\subsection{Suturing Path Planning}

Once the wound is identified, we generate a suitable suturing path in the B-scan plane. We fix the suturing center to reduce the interaction forces between the needle and tissue compared to the moving center suturing approach. To avoid the dead space (Fig.~\ref{fig:woundclose}(a)-(b)), the distance from the left bite point to the deepest point, and the distance from the deepest point to the right exit point should be equal. 
According to standard suturing practice, the suturing depth should exceed $50\%$ of the wound depth. We therefore choose a target suturing depth of $80\%$, from which we can calculate the left bite point and the right exit point. Using the known needle radius, we obtain two solutions for the suture center corresponding to these points in the B-scan plane. One solution lies within the tissue whereas the other is above it. We reject the in-tissue solution because its suturing path does not have the right bite point and the left exit point and use the other solution for planning.

After we obtain the suturing center in 3D, we generate the robot's path in joint space. First, we choose three points in the needle frame (the needle tip $P_{N_{1}}$, the needle body $P_{N_{2}}$, and the needle tail $P_{N_{3}}$, every point is a $3 \times 1$ matrix). These three points determine the suturing needle position and orientation. Second, we uniformly interpolate needle tip $P_{O_{1}}$ milestones along the suturing path in the OCT frame. Because the relationship between these three points is fixed, for each milestone, we calculate the corresponding $P_{O_{2}}$  and $P_{O_{3}}$. Using $T_{OCT}$ from the calibration process, we transfer the milestones from the OCT frame to the world frame ($P_{W_{1}}$, $P_{W_{2}}$ and $P_{W_{3}}$). For each milestone on the suturing path, we use the relationship
\begin{equation}
    \begin{bmatrix}
    P_{W_{1}}&P_{W_{2}}& P_{W_{3}}\\
    1&1&1
    \end{bmatrix} = T_{EE}T_{N}
    \begin{bmatrix}
    P_{N_{1}}&P_{N_{2}}& P_{N_{3}}\\
    1&1&1
    \end{bmatrix}.
    \label{eq:inverse}
\end{equation}
Finally, we use the Klamp't~\cite{Klampt} numerical inverse kinematics solver to compute the robot configuration from the $T_{EE}$ of every milestone. For the first IK solution, the solver is seeded with the robot's current configuration. For the remainder, the solver's is seeded from the last solution.  Collision detection is used to prevent the robot from colliding with the environment or itself.

\begin{figure}[b]
    \centering
    \includegraphics[scale=0.38]{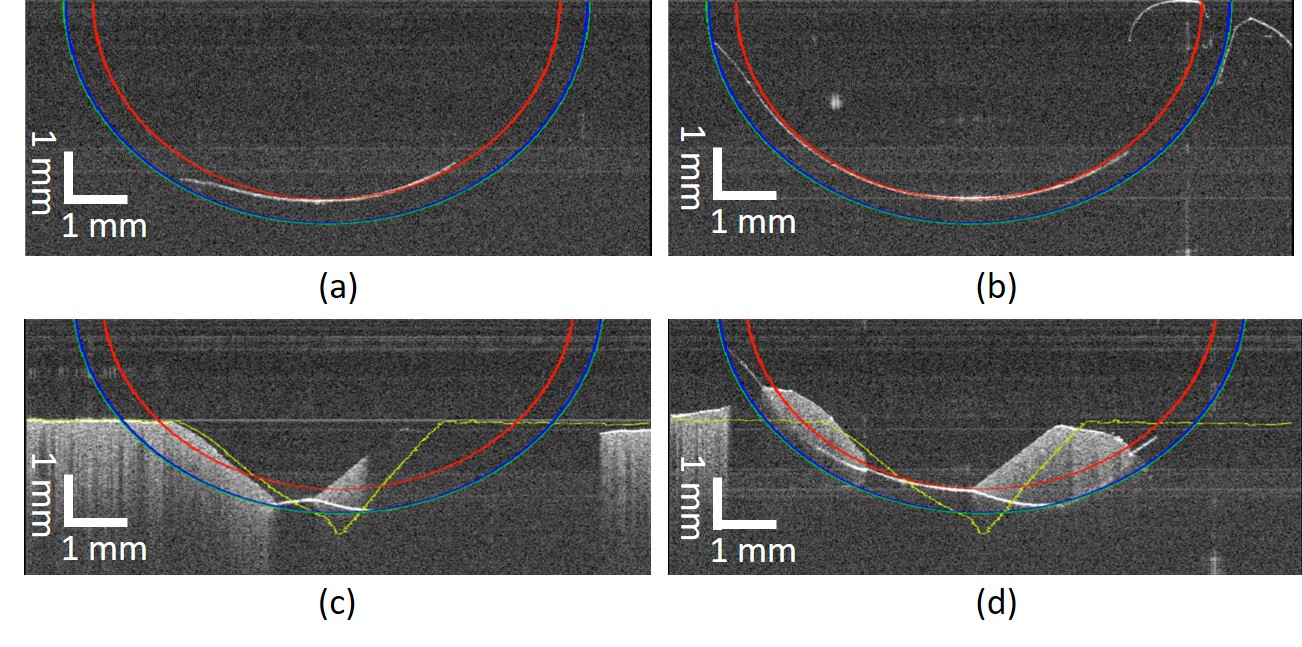}
    \caption{Snapshots of the needle insertion process, showing the OCT B-scan data and the planned path for the needle, showing planned arcs for the inside face (red), outside face (blue), and needle tip (green, almost overlapping with the blue line).  The yellow line is the wound top layer in the undeformed state. In (a) -- (b), the path is executed without the tissue phantom present, showing high accuracy of the calibration process. In (c) -- (d), the needle is inserted into the tissue phantom, showing significant deformation. Note that OCT cannot see the needle outside face path because the laser cannot penetrate metal.  The needle shadows the tissue, and also appears distorted in the tissue due to refraction.}
    \label{fig:insert}
\end{figure}

\section{Experiments and Results}

\subsection{Parameter Selection and Calibration Results}

We evaluated our system in two materials: a silicone rubber tissue phantom (Your Design Medical; Brooklyn, NY) and porcine skin.
Because the scope of this paper does not include needle grasping, we stopped the robot after each full insertion to detach the needle and pull the attached suture through the tract. This allows the tissue to relax to its rest deformation, although the tissue is not perfectly elastic and therefore may not relax completely. The suture path is then imaged using the OCT system.

We performed 25 suturing needle insertions (12 for tissue phantom and 13 for porcine skin). We adjusted the OCT scanner before each attempt to ensure that the wound was within its the field of view. Furthermore, in real surgical settings, the needle will be grasped at a slightly different location every time. To simulate errors caused by grasping, we grasped the needle at the different location on the jaw for each insertion, such that transformations $T_{OCT}$ and $T_{N}$ changed. 
Notably, the wound shape varied across all 25 insertions. The wound width was 2.493\,mm -- 5.110\,mm whereas the wound depth was 1.188\,mm -- 3.082\,mm. The tissue surface was also not necessarily level with tilts of $\pm 9^\circ$ to simulate the real wound environment.

For calibration, we sampled 9 robot configurations to estimate the $T_{OCT}$ and $T_{N}$.
With these 9 robot configurations, the suturing needle was uniformly sampled in the OCT space with different positions and orientations, and we estimated $T_{OCT}$ and $T_{N}$. Comparing the difference between $T_{ICP_{i}}^{-1}$ and $T_{OCT}^{-1}T_{EE_{i}}T_{N}$ for our 25 groups of data, the translation RMSE was 0.075\,mm -- 0.227\,mm, and the rotation RMSE was 0.014\,rad -- 0.090\,rad.

To empirically tune and evaluate wound analysis, we gathered 590 B-scans of different size wounds, and compared our method against manual grading as ground truth.  With the AMAL~\cite{AMAL} parameter $x = 1.02$, AMAL detected all wound top layers successfully.  The denoising Gaussian kernel standard deviation is set to be $8$ pixels, and the gradient threshold to detect the start point of the wound and the end point of the wound is chosen to be $-1.5$ and $+1.5$, respectively.
With these values, our methods successfully detected 582 wounds and keypoints as compared to manual grading (98.6\% success rate).

\subsection{Needle Insertion Results}

\begin{figure}[tbp]
    \centering
    \includegraphics[scale=0.34]{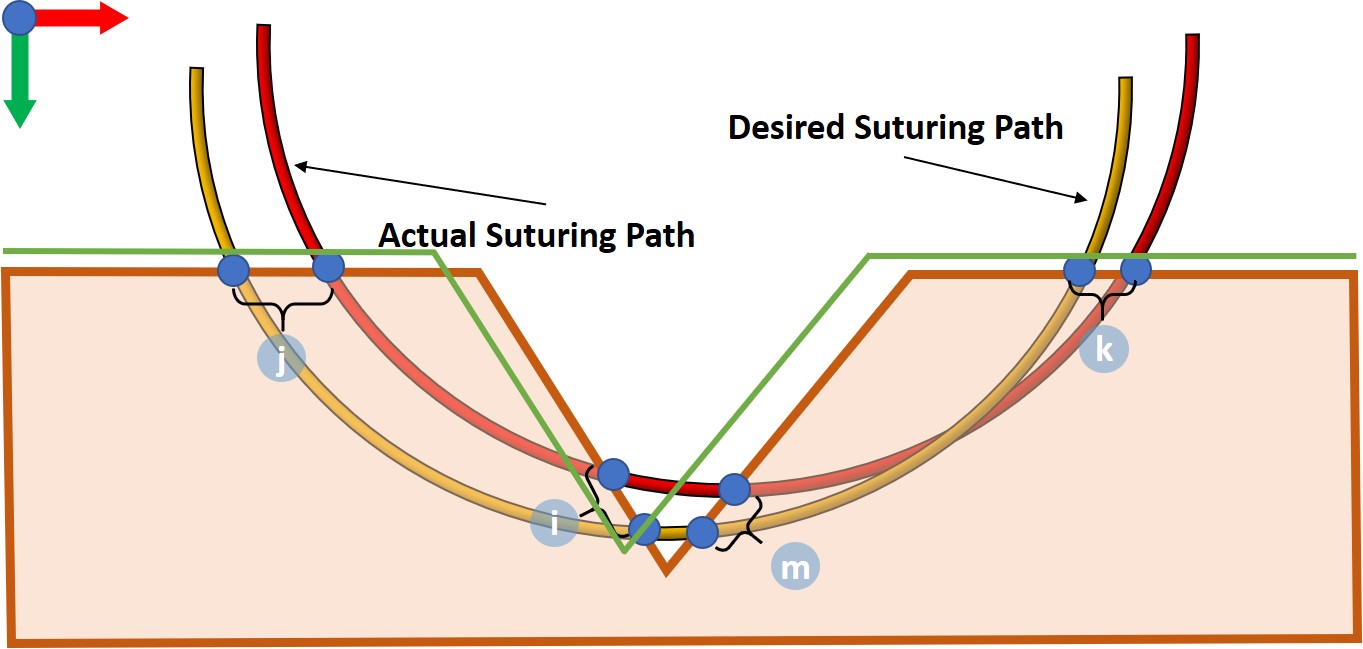}
    \caption{The method used for suture grading. The tissue shape is after the needle pulled out, and the green line is the AMAL result before needle insertions. k) right bite point error. m) right exit point error. i) left bite point error. j) the left exit point error.}
    \label{fig:grade}
\end{figure}

\begin{figure}[tbp]
    \centering
    \includegraphics[scale=0.40]{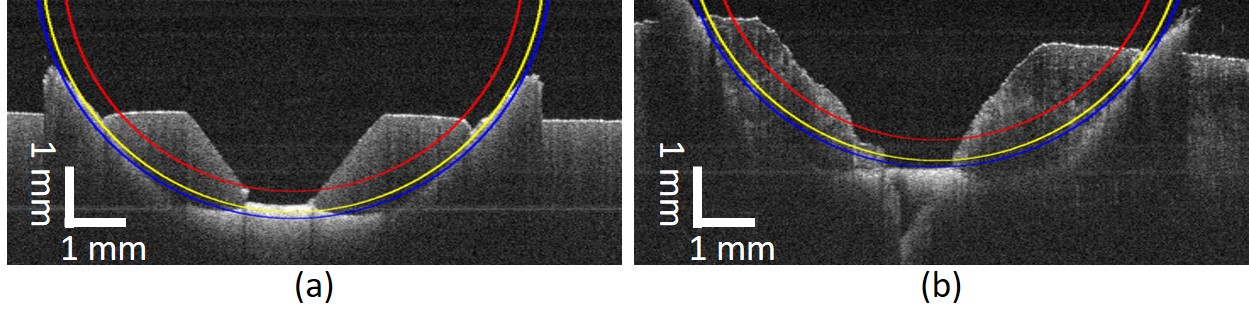}
    \caption{OCT B-scan images of the sutures after the needle has passed through the tissue.  (a) is in the tissue phantom, and (b) is porcine skin.  The planned needle paths are shown for the needle inside face (red), outside face (blue), and suture top layer (yellow).}
    \label{fig:sutu_res}
\end{figure}

We first executed suturing paths in free space to visualize whether the needle follows our planned path.  A representative result is shown in Fig.~\ref{fig:insert}(a)-(b), demonstrating close agreement. 
Next, we performed insertions in the tissue phantom (Fig.~\ref{fig:insert}(c)-(d)).  Because it is difficult to observe the suture path with OCT when the suture passes far below the tissue surface, we graded the accuracy of insertions by calculating the distance between the planned and actual points where the suture intersects the tissue (Fig.~\ref{fig:grade}).  However, due to tissue deformation, it is difficult to judge whether the needle bites and exits at our planned points. Therefore, we manually removed the needle from the needle driver, and pulled the suture through the wound.  Because the friction from the needle is removed and the suture is loose, the wound will approximately return to its resting shape (Fig.~\ref{fig:sutu_res}). By inspection, the sutures' paths are in the planned boundary of the needle. 
The RMSE for each of the four grading points for the tissue phantom and porcine skin are shown in Tables~\ref{tab:phantoms} and~\ref{tab:pork}, respectively.  
The RMSE for each of the grading points and overall was lower than half the width of the needle, and 0.200\,mm averaged over both tissue conditions.

Our final test performed a multi-throw suturing task on the porcine skin (Fig.~\ref{fig:final_suture}). After each insertion, we manually unmounted the needle from the needle driver, pulled the suture through, and then re-mounted the needle for the next throw.  Needle calibration was performed after each throw as usual.  After three throws, we manually closed the wound by pulling on both ends of the suture. OCT images show good alignment of the wound boundary and no dead space.  The problem of autonomous regrasping and wound closure in multi-throw micro-suturing is left for future work.

\begin{table}[t]
    \caption{Tissue Phantoms Suture RMSE (mm, $N=12$)}
    \label{tab:phantoms}
    \begin{center}
    \begin{tabular}{lcccc}
        \toprule
                    & B-scan ($X$) & A-scan ($Y$) & C-scan ($Z$) & 3D    \\
        \midrule
        Right Bite  & 0.081        & 0.059        & 0.142        & 0.174 \\
        Right Exit  & 0.076        & 0.116        & 0.084        & 0.162 \\
        Left Bite   & 0.091        & 0.132        & 0.076        & 0.177 \\
        Left Exit   & 0.151        & 0.077        & 0.091        & 0.192 \\
        \midrule
        Total       & 0.104        & 0.100        & 0.102        & 0.177 \\
        \bottomrule
    \end{tabular}
    \end{center}
\end{table}

\begin{table}[t]
    \caption{Porcine Skins Suture RMSE (mm, $N=13$)}
    \label{tab:pork}
    \begin{center}
    \begin{tabular}{lcccc}
        \toprule
                    & B-scan ($X$) & A-scan ($Y$) & C-scan ($Z$) & 3D    \\
        \midrule
        Right Bite  & 0.070        & 0.049        & 0.111        & 0.140 \\
        Right Exit  & 0.104        & 0.139        & 0.076        & 0.190 \\
        Left Bite   & 0.074        & 0.170        & 0.092        & 0.207 \\
        Left Exit   & 0.269        & 0.097        & 0.107        & 0.305 \\
        \midrule
        Total       & 0.153        & 0.123        & 0.098        & 0.219 \\
        \bottomrule
    \end{tabular}
    \end{center}
\end{table}

\section{CONCLUSION}

We presented a method for OCT-guided calibration and path planning for a suturing needle to be guided autonomously and accurately through soft tissue.  Calibration, wound detection, and path planning are all performed using the input from OCT imaging. Execution performance was graded by the observing the path followed by the suture, demonstrating sub-millimeter errors between the planned and executed entry and exit points, with error magnitude less than half the width of the needle.

In the future, we hope to address some remaining issues in our work. First, the needle calibration error is approximately 20$\times$ larger than the best-reported OCT-based calibration results in the literature, because the curved needle and needle grasping error introduce more challenging initialization and imaging conditions. More advanced image processing techniques may address OCT imaging artifacts such as highlights, refraction, shadowing, and mirror images to perform needle and wound tracking in real time. Solving these issues is a prerequisite for closed-loop guidance during suturing. 






\bibliographystyle{abbrv}
\bibliography{references}

\end{document}